\documentclass[11pt]{article}

\usepackage[preprint]{acl}

\usepackage{times}
\usepackage{latexsym}
\usepackage{booktabs}
\usepackage{multirow}
\usepackage{multicol}

\usepackage[T1]{fontenc}

\usepackage[utf8]{inputenc}

\usepackage{microtype}

\usepackage{inconsolata}

\usepackage{graphicx}
\usepackage{amsmath}
\usepackage{enumitem}
\usepackage{array}
\usepackage{listings}
\usepackage{xcolor}
\usepackage{subcaption}
\usepackage{caption}
\usepackage{xspace}

\newcommand{\figref}[1]{Figure~\ref{#1}}
\newcommand{\tbref}[1]{Table~\ref{#1}}
\newcommand{\secref}[1]{\S\ref{#1}}
\newcommand{\para}[1]{\noindent \textbf{#1}\xspace}

\newif\ifshowcomments
\showcommentsfalse

\ifshowcomments
  \newcommand{\karen}[1]{\textcolor{magenta}{[#1 ---\textsc{kz}]}}
  
  \newcommand{\todo}[1]{\textcolor{red}{[\textsc{todo} --- #1]}}

\else
  \newcommand{\karen}[1]{}
  \newcommand{\todo}[1]{}
\fi

\lstdefinestyle{pythonstyle}{
    language=Python,
    basicstyle=\ttfamily\small,
    keywordstyle=\color{black}\bfseries,
    stringstyle=\color{magenta!75},
    commentstyle=\color{teal!50}\itshape,
    numbers=none,
    backgroundcolor=\color{gray!10},
    showspaces=false,
    showstringspaces=false,
    showtabs=false,
    tabsize=4,
    captionpos=b,
    breaklines=true,
    breakatwhitespace=false,
    frame=single,
    rulecolor=\color{gray!40},
}

\lstnewenvironment{python}[1][]
    {\lstset{style=pythonstyle, literate={\%}{\%}{1}, #1}}
    {}

\lstdefinestyle{bashstyle}{
    language=bash,
    basicstyle=\ttfamily\small,
    keywordstyle=\color{teal}\bfseries,
    stringstyle=\color{orange},
    commentstyle=\color{gray}\itshape,
    numbers=none,
    backgroundcolor=\color{gray!10},
    keywordstyle=\color{cyan}\bfseries,
    stringstyle=\color{orange},
    commentstyle=\color{gray}\itshape,
    showspaces=false,
    showstringspaces=false,
    showtabs=false,
    tabsize=4,
    captionpos=b,
    breaklines=true,
    breakatwhitespace=false,
    frame=single,
    rulecolor=\color{black!60},
}

\lstnewenvironment{bash}[1][]
    {\lstset{style=bashstyle, #1}}
    {}

\makeatletter
\AtBeginDocument{\let\c@lstlisting\c@figure}
\makeatother

\title{\texttt{AutoChecklist}: Composable Pipelines for Checklist Generation \\and Scoring with LLM-as-a-Judge}

\author{Karen Zhou \hspace{1em}
  Chenhao Tan  \vspace{0.5em} \\
  University of Chicago \\
  \texttt{\small\{karenzhou, chenhao\}@uchicago.edu} \\[0.5em]     
  $\vcenter{\hbox{\small\href{https://github.com/ChicagoHAI/AutoChecklist}{\texttt{Code}}}}$%
  \hspace{0.3em}$\vcenter{\hbox{\includegraphics[height=3em]{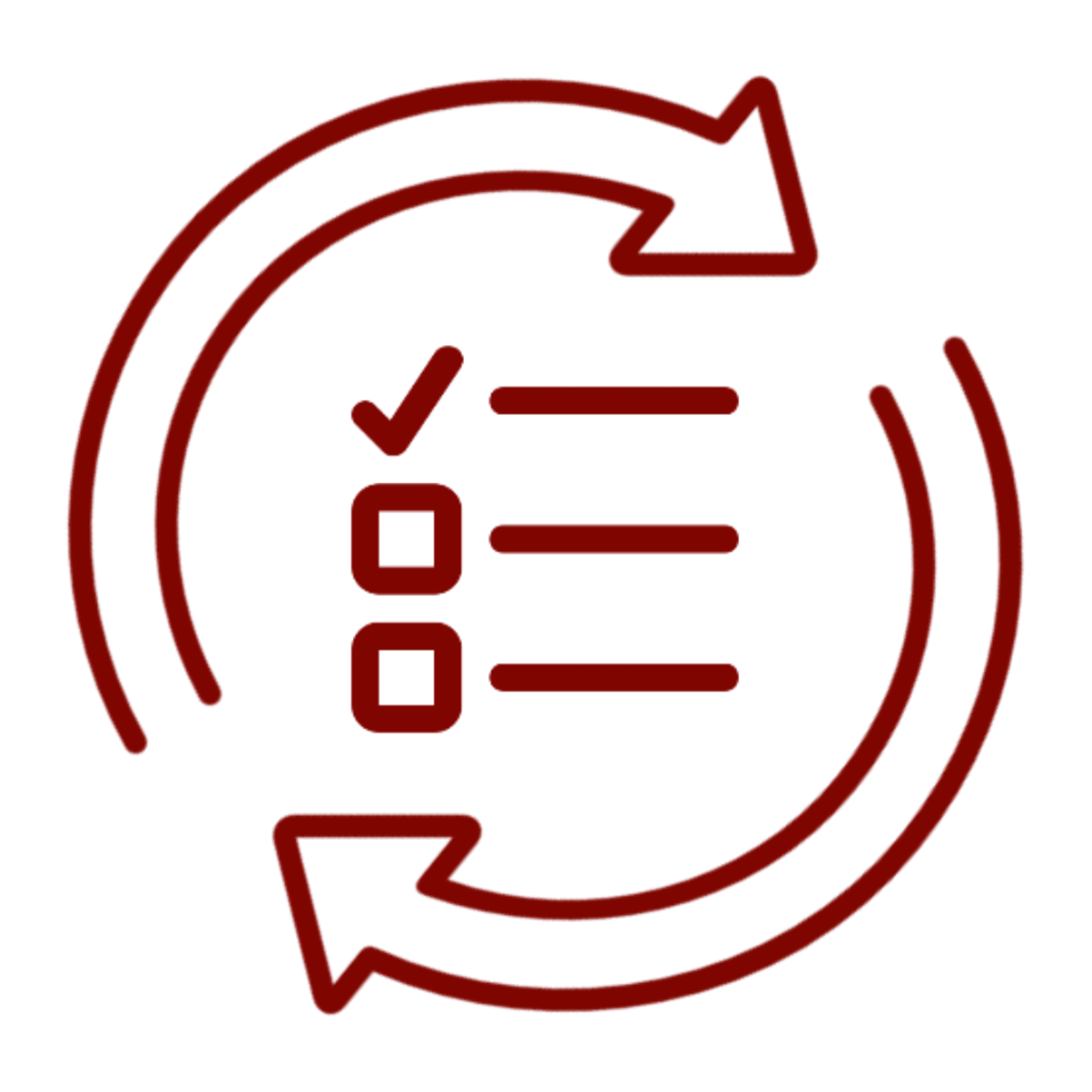}}}$\hspace{0.3em}%
  $\vcenter{\hbox{\small\href{https://autochecklist.github.io}{\texttt{Website}}}}$
  }

\begin{document}
\maketitle
\begin{abstract}
Checklists have emerged as a popular approach for interpretable and fine-grained evaluation, particularly with LLM-as-a-Judge. 
Beyond evaluation, these structured criteria can serve as signals for model alignment, reinforcement learning, and self-correction. To support these use cases, we present \texttt{AutoChecklist}, an open-source library that unifies checklist-based evaluation into composable pipelines.
At its core is a taxonomy of five checklist generation abstractions, each encoding a distinct strategy for deriving evaluation criteria.
A modular \textit{Generator $\rightarrow$ Refiner $\rightarrow$ Scorer} pipeline connects any generator with a unified scorer, and new configurations can be registered via prompt templates alone.
The library ships with ten built-in pipelines implementing published approaches and supports multiple LLM providers (OpenAI, OpenRouter, vLLM).
Beyond the Python API, the library includes a CLI for off-the-shelf evaluation and a web interface for interactive exploration.
Validation experiments confirm that these checklist methods significantly align with human preferences and quality ratings, and a case study on ICLR peer review rebuttals demonstrates flexible domain adaptation.
\texttt{AutoChecklist} is publicly available at \url{https://github.com/ChicagoHAI/AutoChecklist}. 
\end{abstract}

\begin{figure}[t]
    \centering
    \includegraphics[width=\linewidth]{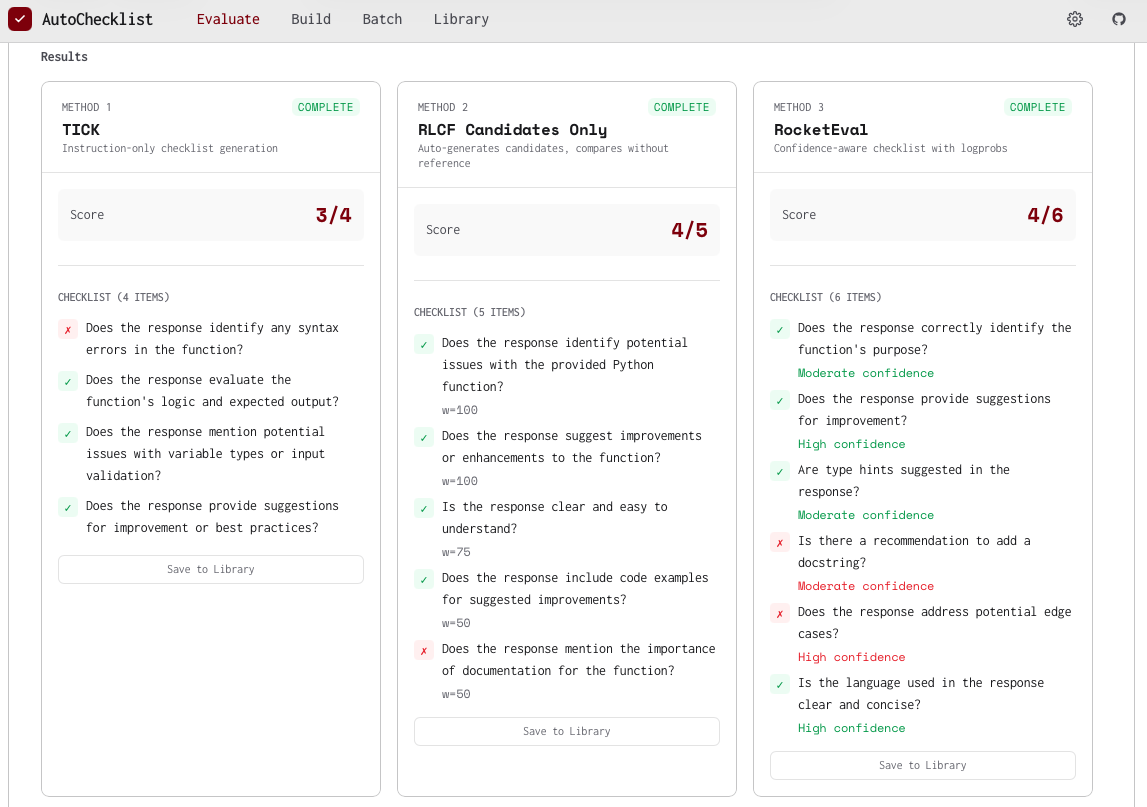}
    \caption{In the \emph{Compare} page of \texttt{AutoChecklist}'s UI, users select methods and enter an input to generate side-by-side checklists, enabling direct comparison of generation strategies on the same task.}
    \label{fig:compare}
\end{figure}

\section{Introduction}

Checklists decompose quality into individually verifiable criteria, with simple yes/no answers that bypass position bias of pairwise comparisons and subjectivity of scalar metrics with LLM-as-a-Judge \citep{ye2024justice}.
As a result, checklists offer interpretable and fine-grained evaluation of text quality, making them an effective tool for evaluation with LLM evaluators \cite{cook2024ticking, viswanathan2025checklists, wei2025rocketeval, zhou-etal-2025-feedback}.

Several checklist generation methods have been proposed in recent work (see \secref{sec:related_work}), each with distinct codebases, prompting strategies, and scoring mechanisms.
These approaches differ along several axes: scope of checklist questions, inclusion of a reference response in checklist generation, unweighted vs.\ weighted scoring,  and question refinement steps, if any.
No existing toolkit provides a unified interface across these methods, making it difficult to compare or extend them for new tasks without significant re-implementation.

To address this, we develop \texttt{AutoChecklist}, an open-source Python library that consolidates common checklist approaches into composable pipelines.
We additionally validate the utility of the library (\secref{sec:validation}) and demonstrate its effectiveness in a new domain (\secref{sec:case-study}).
Our contributions are as follows:
\begin{itemize}[leftmargin=*, itemsep=-0.5em]
    \item A \textbf{taxonomy of five generator abstractions} (direct, contrastive, inductive, deductive, and interactive) that organizes checklist methods by their strategies for producing evaluation criteria.
    \item A framework of \textbf{composable pipelines}: ten built-in configurations implementing published methods, compatible with a unified scorer that consolidates three scoring strategies from the literature. Users can customize these pipelines to new tasks via Markdown prompt templates alone.
    \item A \textbf{\texttt{pip}-installable Python package} with multiple usage modes, including a CLI for off-the-shelf evaluation with pre-defined pipelines.
    \item \textbf{Multi-provider LLM backend} support for OpenAI, OpenRouter, and vLLM (including \texttt{VLLMOfflineClient} for local GPU inference), and automatic handling of structured output.
    \item A \textbf{locally hosted user interface} supporting most package functionality, allowing for interactive prompt customization, pipeline configuration and comparison (e.g., \figref{fig:compare}), and batch evaluation. 
\end{itemize}

\section{Related Work}
\label{sec:related_work}

\para{Checklist Generation \& Evaluation.}
Checklist-based evaluation with LLM-as-a-Judge has been leveraged for a variety of tasks and benchmarks \citep{lin2024wildbenchbenchmarkingllmschallenging,que2024hellobenchevaluatinglongtext,qin-etal-2024-infobench}.
A common evaluation setup is to generate a checklist per input-output pair, such as for a user query and LLM response. Checklist generation can either use zero- to few-shot prompting to generate yes/no questions from the input alone, or by incorporating a reference and/or alternative response samples \cite{cook2024ticking,wei2025rocketeval, viswanathan2025checklists}. 
Other methods aim to derive a shared checklist from corpus-wide signals. These signals are either pre-defined by humans and then processed into granular criteria, or extracted from brainstorming or feedback items into structured checklists \cite{lee-etal-2025-checkeval,chu2025think,zhou-etal-2025-feedback}. These methods often involve multiple steps to refine the initial checklist questions, such as through deduplication, filtering, or elaboration.

Despite their differences, all of these methods share a common structure: generate criteria from an input, optionally refine them, then score a target response.
\texttt{AutoChecklist} builds on this shared structure.
We identify five generator abstractions, each encoding a distinct reasoning strategy, and factor them into interchangeable generators, refiners, and scorers within a single library.
To our knowledge, this is the first unified toolkit for LLM-based checklist generation.

\vspace{0.1em}
\para{Utility of Checklists.}
Beyond evaluation, structured criteria serve as signals for alignment \citep{viswanathan2025checklists, siro2026gereval}, reinforcement learning \citep{yang2026healthscore}, and self-correction \citep{wan2026deepverifier}.
However, \citet{furuhashi2025checklists} find that checklists are not consistently useful and highlight the need to more clearly define objective evaluation criteria.
\texttt{AutoChecklist} aims to lower the barrier for both human and LLM judges to define and select such criteria.

\vspace{0.1em}
\para{Unified Frameworks.} 
\href{https://github.com/quotient-ai/judges}{judges} is a similar library to use and create LLM-as-a-Judge evaluators, curating a set of LLM evaluators in a low-friction format that can be used off-the-shelf or customized. Numerous other open-source packages exist for evaluating and testing LLM systems (\href{https://github.com/confident-ai/deepeval}{DeepEval}, \href{https://github.com/vibrantlabsai/ragas}{RAGAS}, \href{https://github.com/openai/evals}{OpenAI Evals}). 
While the generated checklists from \texttt{AutoChecklist} are primarily used with LLM-as-a-Judge, their usefulness is not limited to automatic evaluation; there is also potential for such checklists to help further align human annotators in subjective tasks (e.g., essay grading, clinical note assessment). Furthermore, \texttt{AutoChecklist} features unified criteria \textit{generation}; most other libraries focus on scoring.

\begin{table*}[t]
    \centering
    \small
    \begin{tabular}{p{6em} p{3em} p{23em} p{11em}}
    \toprule
        Generator &  Level &  Strategy  & Pipelines \\
        \cmidrule(lr){1-4}
        \textsc{Direct} & Instance & Single-step generation: given the input (and optionally a reference), directly prompt the LLM to produce a checklist &  TICK \citep{cook2024ticking}, RLCF Direct \citep{viswanathan2025checklists}, RocketEval \citep{wei2025rocketeval}\\
        \textsc{Contrastive} & Instance & Counterfactual reasoning: generate candidate responses of varying quality, then derive criteria by contrasting good vs.\ bad responses & RLCF Candidate-based \citep{viswanathan2025checklists}, CRG \citep{liu2026openrubricsscalablesyntheticrubric} \\
        \textsc{Inductive} & Corpus & Bottom-up induction: convert reviewer feedback or observations into general evaluation criteria, with built-in deduplication and selection & \citet{zhou-etal-2025-feedback} \\
        \textsc{Deductive} & Corpus & Top-down decomposition: convert expert-defined evaluation dimensions into specific checklist questions, with optional augmentation  &  \citet{lee-etal-2025-checkeval}\\
        \textsc{Interactive} & Corpus & Starting with defined dimensions, extract criteria from human and LLM think-aloud evaluation sessions via a multi-stage pipeline of clustering and question generation  &  \citet{chu2025think} \\
    \bottomrule
    \end{tabular}
    \caption{Five checklist generation abstractions in \texttt{AutoChecklist}. Each represents a distinct strategy for deriving evaluation criteria. The rightmost column lists built-in pipelines that instantiate each generator with paper-specific prompt templates and scoring strategies.}
    \label{tab:generators}
\end{table*}

\section{System Details}
\label{sec:details}

A \textit{checklist} $C = \{q_1, \ldots, q_n\}$ is a set of yes/no questions designed to assess a target text.
Each question $q_i$ may carry an optional importance weight $w_i \in [0, 100]$.
Given a target response, a \textit{scorer} assigns each item an answer $a_i \in \{\textsc{yes}, \textsc{no}\}$ and optionally a confidence $c_i \in [0, 1]$.
The \textit{pass rate} for a single input is $|\{a_i = \textsc{yes}\}| / |C|$, or $\sum w_i \cdot \mathbf{1}[a_i{=}\textsc{yes}] \,/\, \sum w_i$ when importance weights are available.

Since different methods are originally implemented with specific tasks in mind (e.g., evaluating LLM instruction-following or quality of a summary), we further define the following abstractions: a checklist \textit{input} is the instruction, query, or task being evaluated (e.g., ``Write a haiku about autumn'' or ``Summarize: \{article\}''); the checklist or evaluation \textit{target} is the output being scored against the checklist (e.g., the haiku or summary).

\begin{figure}
    \centering
    \includegraphics[width=\linewidth]{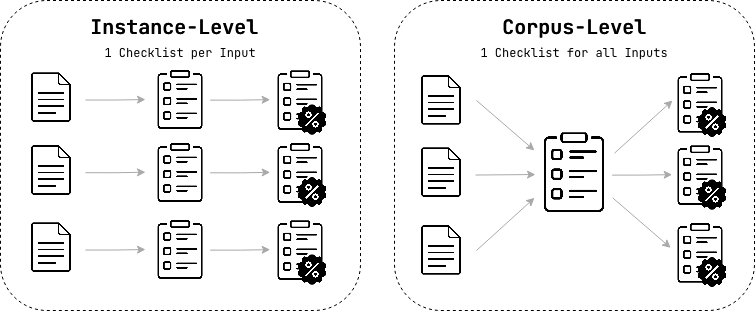}
    \caption{{\it Instance-level} generators produce one checklist per input;  criteria are tailored to each specific task (target). {\it Corpus-level} generators produce one checklist for an entire dataset of targets, capturing general quality patterns derived from higher-level signals.}
    \label{fig:instance_vs_corpus}
\end{figure}

\vspace{0.1em}
\para{Design and Architecture.}
As depicted in \figref{fig:instance_vs_corpus}, generators fall into two categories.
\textit{Instance-level generators} produce a checklist per input (e.g., per prompt-response pair).
\texttt{DirectGenerator} (\textsc{Direct}) implements direct inference: a single LLM call that converts the input into checklist questions.
\texttt{ContrastiveGenerator} (\textsc{Contrastive}) implements counterfactual reasoning; it first generates candidate responses of varying quality (usually from weaker models), then derives discriminative criteria by contrasting them. If the user provides preference pairs, the generator can skip candidate generation and directly derive criteria by contrasting chosen and rejected responses.
\textit{Corpus-level generators} derive a shared checklist from corpus-wide signals.
\texttt{InductiveGenerator} (\textsc{Inductive}) creates checklists ``bottom-up'' from unstructured observations, such as user feedback or reviews.
\texttt{DeductiveGenerator} (\textsc{Deductive}) performs top-down decomposition from expert-defined evaluation dimensions.
\texttt{InteractiveGenerator} (\textsc{Interactive}) extracts criteria from simulated think-aloud protocols, wherein participants verbalize their thoughts during tasks. \tbref{tab:generators} shows how several papers ultimately share similar underlying steps for checklist generation.

\begin{table}[t]
\centering
\small
\begin{tabular}{p{7em} p{6em} p{3em} p {2em}}
\toprule
\textbf{Pipeline} & \textbf{Generator} & \textbf{Score} & \textbf{Ref?} \\
\cmidrule(lr){1-4}
\texttt{tick} & \textsc{Direct}  & pass& No \\
\texttt{rocketeval} & \textsc{Direct}  & norm. & Yes \\
\texttt{rlcf\_direct} & \textsc{Direct}  & weight. & Yes \\
\texttt{rlcf\_cand.} & \textsc{Contrastive} & weight. & Yes \\
\texttt{rlcf\_cand.\_only} & \textsc{Contrastive}  & weight. & No \\
\texttt{or\_pairwise} & \textsc{Contrastive} & pass & No \\
\texttt{or\_listwise} & \textsc{Contrastive} & pass & No \\
\texttt{checkeval} & \textsc{Deductive} & pass  & --- \\
\texttt{interacteval} & \textsc{Interactive} & pass & ---\\
\texttt{feedback} & \textsc{Inductive} & pass & ---\\
\bottomrule
\end{tabular}
\caption{Built-in pipelines. Each pipeline pairs a generator with a default \texttt{ChecklistScorer} configuration and paper-specific prompt templates. Score metrics: pass rate, normalized, or weighted (see \secref{sec:details}).}
\label{tab:pipelines}
\end{table}

\vspace{0.1em}
\para{Refiners.}
Refiners are optional post-processing steps composed in sequence before scoring.
The \texttt{Deduplicator} merges semantically redundant questions via embedding similarity and LLM-based cluster merging.
The \texttt{Tagger} applies additional filtering of items by a specified quality (e.g., generality, specificity).
The \texttt{UnitTester} validates that each item is ``LLM enforceable'' \citep{zhou-etal-2025-feedback}.
The \texttt{Selector} uses beam search to optimize an objective function against checklist length.
For most of the corpus-level generators, refinement steps are built into the generator classes; future work can refactor them into more modular steps.

\vspace{0.1em}
\para{Scorer.}
A unified \texttt{ChecklistScorer} class consolidates scoring strategies from three different papers into a single configurable interface.
It operates in two modes: \textit{batch} mode scores all checklist questions in a single LLM call, while \textit{item} mode scores one question per call. It can configure use of chain-of-thought reasoning (\texttt{capture\_reasoning}) and logprob-derived confidence scores (\texttt{use\_logprobs}).
The scorer computes three aggregate metrics: \texttt{pass\_rate} (fraction of YES answers), \texttt{weighted\_score} ($\sum w_i \cdot \mathbf{1}[a_i{=}\textsc{yes}] / \sum w_i$, using importance weights as in \citet{viswanathan2025checklists}), and \texttt{normalized\_score} (calibrated from logprob confidence as in \citet{wei2025rocketeval}).
A \texttt{primary\_metric} parameter designates which metric the \texttt{primary\_score} attribute returns; each built-in pipeline sets this to match its original paper's scoring method.
For a corpus of targets, both macro-average scores and Decomposed Requirements Following Ratio (DRFR), i.e., micro-average pass rate \citep{qin-etal-2024-infobench}, are calculated.

\vspace{0.1em}
\para{Pipelines.} 
Each stage in a pipeline is its own unit; any generator can be paired with any scorer, and refiners compose in sequence.
For instance, a user can generate checklists with the \textsc{Interactive} strategy but score them with the normalized scorer, a previously unsupported combination.

\tbref{tab:pipelines} shows how the five generator abstractions are instantiated as ten built-in pipelines, i.e., pre-set configurations.
Each row pairs a generator with a default scorer configuration and paper-specific prompt templates, but any generator can be combined with any scorer. 
Users can add new pipelines by registering custom prompt templates, without modifying library code. Base classes are also available for ease of developing more advanced custom components.

\section{Deployment and Usage}

\texttt{AutoChecklist} is \verb|pip|-installable and supports three usage modes with increasing customization: a CLI for off-the-shelf evaluation, a web interface for interactive exploration and prompt iteration, and a Python API for full pipeline control.
The library supports multiple LLM backends via a shared \texttt{LLMClient} protocol: OpenAI API, OpenRouter, and vLLM (both as an online server and via \texttt{VLLMOfflineClient} for local GPU inference).
Structured output is handled via provider-enforced JSON schema (\texttt{response\_format}), or appended format instructions and parsing as a fallback.

\begin{bash}[caption={CLI: end-to-end evaluation with a built-in pipeline.}]
autochecklist run --pipeline tick \
  --data eval_data.jsonl \
  --output results.jsonl \
  --generator-model openai/gpt-5-mini \
  --scorer-model openai/gpt-5-mini
\end{bash}

\para{Command-Line Interface.}
The CLI provides three subcommands for use with registered components: \texttt{run} (generate checklists and score in one step), \texttt{generate} (produce checklists without scoring), and \texttt{score} (score targets against a pre-existing checklist).
A fourth subcommand, \texttt{list}, discovers available generators, scorers, and refiners.
Users specify a built-in pipeline (e.g., \texttt{-{}-pipeline tick}), a custom generator prompt file (\texttt{-{}-generator-prompt my\_eval.md}), or a saved pipeline config JSON.
Runs are resumable after interruption, with incremental saving of results.

\begin{figure}[t]
    \centering
    \includegraphics[width=\linewidth]{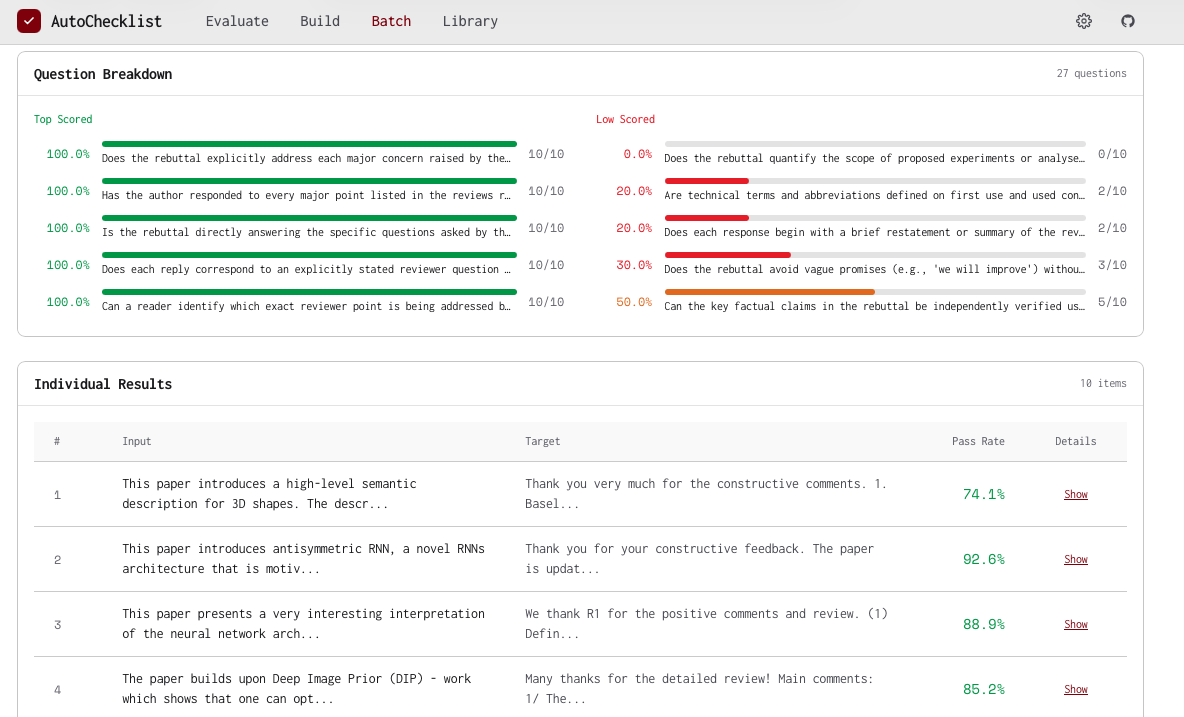}
    \caption{The Batch results page showing per-item score breakdowns and aggregate statistics for a dataset evaluation run.}
    \label{fig:batch_results}
\end{figure}

\vspace{0.1em}
\para{User Interface.}
A locally hosted interface built with FastAPI  and Next.js provides interactive access to most library functionality without writing code, easily launched with \texttt{autochecklist ui}.

The main page \textbf{Evaluate} offers three tabs.
\emph{Custom Eval} provides full prompt editing: users choose a generator class, edit generator and scorer prompts in tabbed editors, load from saved prompt templates, select an output format and scorer type, then run an evaluation and view checklist and score results.
The \emph{Compare} tab enables side-by-side comparison of built-in methods and custom pipelines on the same input, with results displayed as horizontally scrollable method cards (see \figref{fig:compare}). \emph{Reference} includes descriptions for all generators, scorer options, and pipelines. The \textbf{Build} interface supports \texttt{Deductive} corpus-level checklist generation for downstream use.

Configurations from these workflows can be saved to the \textbf{Library}, which stores \emph{Checklists} for saved checklists (clicking in allows for edits), \emph{Prompt Templates} for reusable prompts with placeholders, and \emph{Pipelines} for saving complete evaluation configurations that bundle generator class, prompts, scorer settings, and output format.

To process more data samples at once, the \textbf{Batch} page supports CSV/JSON data upload, evaluation mode with saved Library items, progress tracking, and a results page with basic aggregate statistics and per-item score breakdowns (such as in \figref{fig:batch_results}). There is also the option to export the experiment in Python code to run the same pipeline through a script, which is preferred for very large datasets and for more fine-grained customization.

\vspace{0.1em}
\para{Python API.}
The Python API offers full control over the pipeline, ideal for large-scale checklist evaluation with finalized generator and scorer configurations.
The \texttt{pipeline()} factory is the recommended entry point: it resolves a pipeline name to a pre-configured generator and scorer, accepts optional refiners, and returns a callable that generates and scores in one step.
For finer-grained control, \texttt{ChecklistPipeline} allows explicit composition of independently configured generators, refiners, and scorers.
Components can also be used outside of any pipeline (e.g., calling a generator and scorer directly).
Batch results provide both macro-average scores and Decomposed Requirements Following Ratio (DRFR, i.e., a micro-average pass rate) \citep{qin-etal-2024-infobench}, and can be exported as DataFrames or JSONL files.
Users can register custom pipelines via \texttt{register\_custom\_pipeline()} and share configurations as JSON files, making new evaluation setups reusable without modifying library code.
See Appendix~\ref{sec:appendix} for code examples.

\begin{table}[t]
\centering
\small
\setlength{\tabcolsep}{2pt}
\begin{tabular}{lllcrr}
\toprule
Pipeline  & W/L/T & Mean $\Delta$ & Cohen's $d$ & $t$-test $p$ & Wilcoxon $p$ \\
\cmidrule(lr){1-6}
\texttt{tick} & \textbf{75}/10/15 & \textbf{+0.286} & \textbf{0.919} & 6.3e-15$$ & 4.3e-12$$ \\
\texttt{rlcf} &  70/16/14 & +0.228 & 0.785 & 4.9e-12$$ & 1.1e-10$$ \\
\bottomrule
\end{tabular}
\caption{Preference discrimination on RewardBench pairs using instance-level pipelines. Win = chosen scores higher than rejected, and W/L/T = ratio of Win/Loss/Tie scores.
}
\label{tab:rewardbench-comparison}
\end{table}

\section{Validation}
\label{sec:validation}

We validate that our pipeline implementations produce useful and effective checklists for two benchmarks: RewardBench \cite{lambert-etal-2025-rewardbench} for instance-level methods and SummEval \cite{fabbri-etal-2021-summeval} for corpus-level methods.
All experiments use \verb|gpt-5-mini| as the checklist generator and \verb|gpt-5| as the scorer.

\vspace{0.1em}
\para{Instance-Level.}
We test whether instance-level checklist scores can discriminate between preferred and rejected responses.
From RewardBench, we sample 100 preference pairs and generate a checklist for each pair's input prompt using the \texttt{tick} (\textsc{Direct}, reference-free) and \texttt{rlcf\_candidate\_only} (\textsc{Contrastive}, reference-free) pipelines.
Each method scores both the chosen and rejected response, and we measure win ratio (fraction where the chosen response scores higher), mean score delta, and effect size.

\tbref{tab:rewardbench-comparison} shows that both pipelines significantly discriminate preferred responses.
\texttt{tick} achieves a 75\% win rate with a large effect size ($d = 0.919$, $p < .001$), and \texttt{rlcf\_candidate\_only} achieves 70\% ($d = 0.785$, $p < .001$).
These results confirm that the library's instance-level pipelines produce checklist criteria that align with human preference judgments of quality.

\begin{table}[t]
\centering
\small
\setlength{\tabcolsep}{3pt}
\begin{tabular}{llcccc}
\toprule
Pipeline & Dimension & $\rho_s$ & $r$ & $\tau$ & MAE \\
\cmidrule(lr){1-6}
\texttt{checkeval} & Coherence & 0.723$$ & 0.686 & 0.571 & 1.104 \\
 & Consistency & 0.756$$ & 0.709 & 0.597 & 0.925 \\
 & Fluency & \textbf{0.819}$$ & \textbf{0.814} & \textbf{0.669} & \textbf{0.829} \\
 & Relevance & 0.571$$ & 0.540 & 0.430 & \textbf{0.719} \\
\cmidrule(lr){1-6}
\texttt{interacteval}  & Coherence & \textbf{0.755}$$ & \textbf{0.763} & \textbf{0.586} & \textbf{0.703} \\
 & Consistency & \textbf{0.835}$$ & \textbf{0.835} & \textbf{0.699} & \textbf{0.565} \\
 & Fluency & 0.670$$ & 0.713 & 0.534 & 0.980 \\
 & Relevance & \textbf{0.612}$$ & \textbf{0.610} & \textbf{0.460} & 0.910 \\
\bottomrule
\end{tabular}
\caption{Corpus-level checklist validation on SummEval. $\rho_s$: Spearman, $r$: Pearson, $\tau$: Kendall; MAE on rescaled 1--5 scores. All $\rho_s$, $r$,  $\tau$ are significant ($p<.001$).}
\label{tab:summeval-correlations}
\end{table}

\vspace{0.1em}
\para{Corpus-Level.}
We validate corpus-level methods by measuring correlation with expert quality judgments on the SummEval benchmark, which provides human scores on a 1--5 scale across four dimensions: coherence, consistency, fluency, and relevance.
We generate corpus-level checklists using \texttt{checkeval} (\textsc{Deductive}) and \texttt{interacteval} (\textsc{Interactive}), then score 100 summaries against each checklist.
We map checklist scores to the 1--5 scale (pass rate $\times$ 4 + 1).

Both methods achieve strong correlations across all four dimensions (\tbref{tab:summeval-correlations}).
\texttt{interacteval} achieves the highest single-dimension correlation on consistency ($\rho = 0.835$, $p < .001$) and outperforms \texttt{checkeval} on three of four dimensions.
\texttt{checkeval} achieves the strongest correlation on fluency ($\rho = 0.819$, $p < .001$) and the lowest Mean Absolute Error (MAE) on relevance (0.719).
All correlations are significant at $p < .001$.
These results confirm that the library's corpus-level pipelines produce checklists whose pass rates align with expert judgments.

\section{Case Study: Author Rebuttal Checklists}
\label{sec:case-study}

To demonstrate how \texttt{AutoChecklist}'s composable design enables application to new domains, we apply checklist-based evaluation to peer review rebuttals, a setting where no existing checklist method has been tested. 
The adaptation required only prompt modifications and the library's existing generator and scorer infrastructure.

Review-rebuttal dynamics of peer review have been studied extensively, uncovering their effects on final scores and successful rebuttal strategies \citep{gao-etal-2019-rebuttal,10.1145/3584664,he2026rebuttalagent,ma2026paper2rebuttalmultiagentframeworktransparent}. 
While there are efforts to evaluate review quality and usefulness \citep{ryu2025reviewscoremisinformedpeerreview,sadallah-etal-2025-good}, no prior study has applied checklist-based evaluation to automatically assess {\it rebuttal} quality. 
Checklists offer a natural fit: each item can target a specific concern raised by the reviewer, yielding interpretable, fine-grained assessment of rebuttal responses.

\para{Setup.}
We collect 100 review--rebuttal pairs via the OpenReview API and evaluate them with four generator configurations spanning both instance-level and corpus-level approaches. We select pairs from ICLR 2019, the latest year with post-rebuttal rating change data.

We register custom generator prompts specific to the peer review context. Additionally,
\textsc{Deductive} uses three dimensions from \citet{ma2026paper2rebuttalmultiagentframeworktransparent}---Relevance, Argumentation Quality, and Communication Quality---each with sub-dimensions (e.g., Semantic Alignment, Evidence Support, Constructiveness).
\textsc{Inductive} takes 1,000 labeled reviewer sentences related to ``weaknesses'' as input, generates candidate checklist items that capture common patterns of concern across the corpus, and then refines the checklist with deduplication, tagging, and optimal selection.

\vspace{0.1em}
\para{Results.}
We evaluate each pipeline's checklist pass rates against three external signals of rebuttal quality: (1) \textsc{Rating}: Spearman correlation between pass rate and the reviewer's numeric rating; (2) \textsc{Acceptance}: rank-biserial effect size between accepted and rejected papers; and (3) \textsc{Rating Change}: point-biserial correlation and ROC-AUC for predicting whether the reviewer changed their rating post-rebuttal.

\begin{table}[t]
\centering
\small
\setlength{\tabcolsep}{2pt}
\begin{tabular}{l  c  cc  cc}
\toprule
& \multicolumn{1}{c}{\textsc{Rating}} & \multicolumn{2}{c}{\textsc{Acceptance}} & \multicolumn{2}{c}{\textsc{Rating Change}} \\
\cmidrule(lr){2-2} \cmidrule(lr){3-4} \cmidrule(lr){5-6}
Generator & $r_s$ & $|d|$ & $\Delta\bar{x}$ & $r_{pb}$ & AUC \\
\cmidrule(lr){1-6}
\textsc{Direct} & 0.253$^{*}$ & \textbf{0.296}$^{*}$ & \textbf{+0.124} & 0.143 & 0.587 \\
\textsc{Contrastive} & 0.219$^{*}$ & 0.270$^{*}$ & +0.094 & 0.084 & 0.531 \\
\cmidrule(lr){1-6}
\textsc{Deductive} & \textbf{0.267}$^{**}$ & 0.282$^{*}$ & +0.094 & \textbf{0.242}$^{*}$ & \textbf{0.668} \\
\textsc{Inductive} & 0.192 & 0.240$^{*}$ & +0.068 & 0.172 & 0.630 \\
\bottomrule
\end{tabular}
\caption{Evaluation of checklist-based rebuttal scoring on ICLR 2019. \textsc{Rating}: Spearman $r_s$ between pass rate and reviewer rating. \textsc{Acceptance}: rank-biserial effect size $|d|$ and pass rate difference $\Delta\bar{x} = \bar{x}_{acc} - \bar{x}_{rej}$. \textsc{Rating Change}: point-biserial $r_{pb}$ and ROC-AUC for predicting whether the reviewer changed their rating post-rebuttal. $^{*}p<.05$,\ $^{**}p<.01$,\ $^{***}p<.001$.}
\label{tab:summary-stats}
\end{table}

\tbref{tab:summary-stats} reveals the patterns across generator types.
\textsc{Direct} shows the strongest acceptance discrimination 
and moderate rating correlation. 
\textsc{Contrastive} shows similar patterns with slightly lower effect sizes. 
\textsc{Deductive} achieves the highest rating correlation,
and 
both \textsc{Deductive} and \textsc{Inductive} are the only methods with significant rating-change prediction, i.e., corpus-level checklists may better capture the signals in rebuttals that convince reviewers to update their score.

This case study illustrates two features of the library's design.
First, the custom prompt API allows adapting any pipeline to a new domain without modifying library code, as the existing generator and scorer infrastructure is reusable with custom prompts.
Second, the composable pipeline makes it straightforward to compare instance-level and corpus-level approaches on the same data, revealing their complementary strengths.

\section{Conclusion and Future Work}

We present \texttt{AutoChecklist}, a composable Python library that unifies LLM-based checklist evaluation.
Beyond consolidating existing methods, the library contributes a taxonomy of five generator abstractions that organizes the design space of checklist generation by reasoning strategy, as well as a unified \texttt{ChecklistScorer} that consolidates three scoring strategies.
These utilities, combined with the composable \textit{Generator $\rightarrow$ Refiner $\rightarrow$ Scorer} architecture, enable systematic comparison and rapid extension of checklist evaluation methods.

Validation on RewardBench and SummEval confirms that library-generated checklists align with human judgments.
The peer review rebuttal case study (\secref{sec:case-study}) demonstrates that adapting to a new domain requires only new custom prompt templates and no library code changes.
Future directions include integrating new checklist methods as they emerge, supporting human-in-the-loop generation and refinement for corpus-level checklists, and providing more library features through the UI.

\para{Limitations.} 
We aim to faithfully implement the built-in pipelines based on their papers and code, but we make some changes for standardization (e.g., enforcing JSON structured outputs) and generalization (e.g., modifying original prompts). 
We provide default prompts calibrated to the original papers, but users adapting to new domains should expect to iterate on prompt templates.
Finally, there is not yet full end-to-end support for corpus-level pipelines that require multi-stage human-in-the-loop workflows (e.g., \textsc{Inductive}'s refinement pipeline and an interface to collect human and LLM data for \textsc{Interactive}'s input attributes).

\section*{Ethical Considerations}
\karen{not counted in page limit}
Automatic checklist-based evaluation is imperfect, especially with multiple LLM-based steps. We caution that checklist scores should not replace human judgment in high-stakes settings without careful validation.

The library is released under the Apache 2.0 license. Built-in pipelines are adapted based on publicly available papers and code. Validation experiments use open-source benchmarks (RewardBench, SummEval) and publicly accessible OpenReview data. No private or personally identifiable data is used. Users do not expose API keys or data through using the package or UI.

\section*{Acknowledgments}
We thank the Chicago Human+AI Lab for helpful discussions and feedback during this work.

\bibliography{custom}


\appendix

\onecolumn

\section{Additional Figures}

\begin{figure}[h]
    \centering
    \begin{subfigure}[t]{0.45\linewidth}
    \centering
        \includegraphics[width=\linewidth, height=7cm, keepaspectratio]{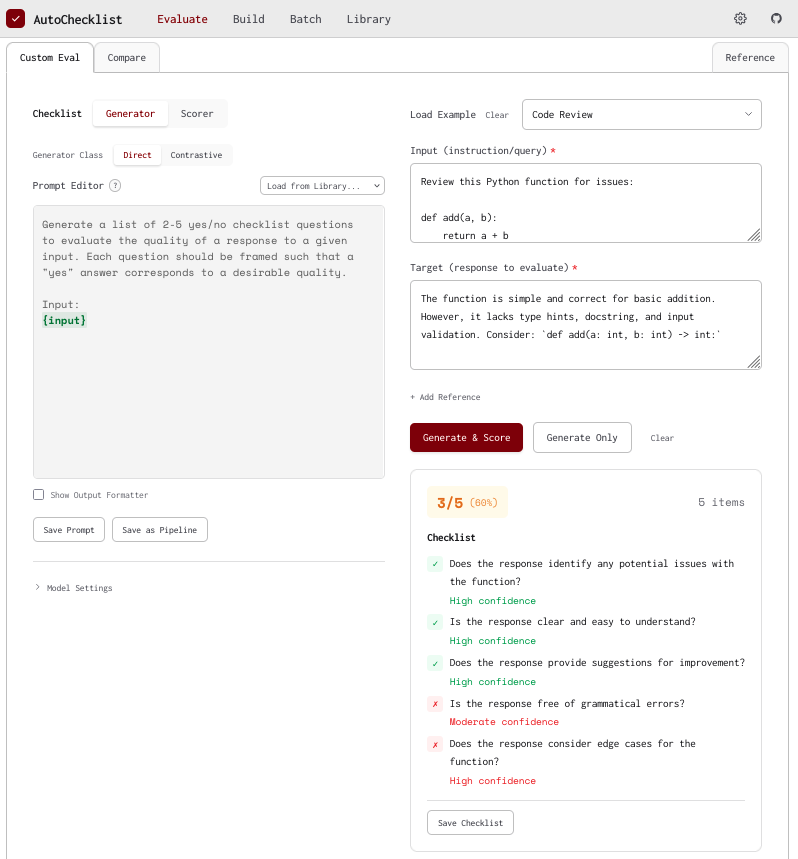}
        \vspace{1pt}
        \caption{The {\it Custom Eval} tab, where users edit generator and scorer prompts, select an output format, and run single-input evaluations.}
        \label{fig:custom_eval}
    \end{subfigure}
    \hfill
    \begin{subfigure}[t]{0.45\linewidth}
    \centering  
        \includegraphics[width=\linewidth, height=7cm, keepaspectratio]{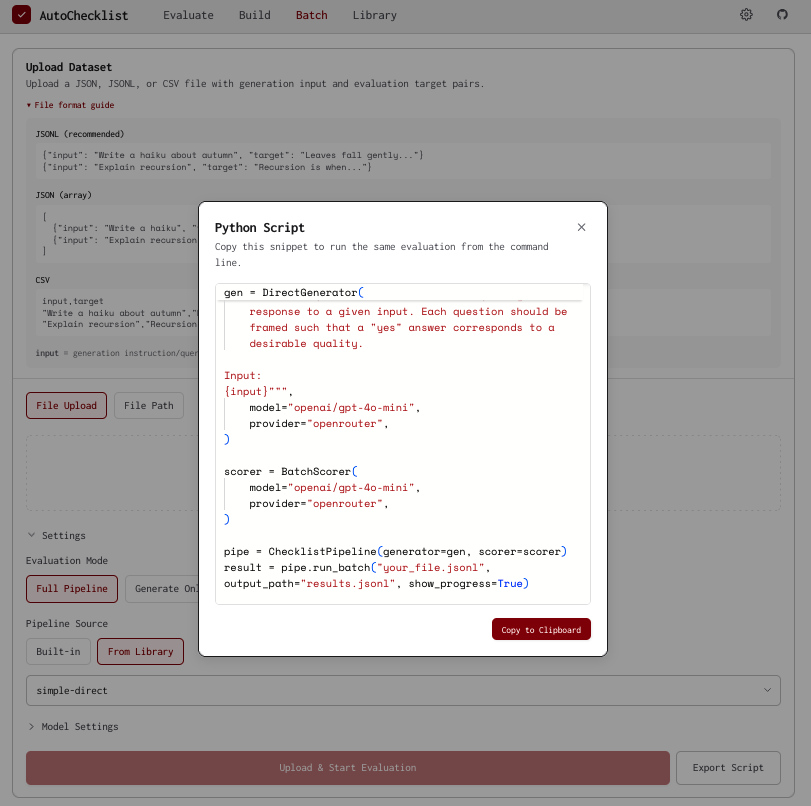}
        \vspace{1pt}
        \caption{The {\it Batch} page, where users can use built-in and custom pipelines to generate checklists and/or score a dataset, along with an option to export the pipeline as a script.}
        \label{fig:batch}
    \end{subfigure}
    \caption{Additional UI pages to support prompt iteration and dataset evaluation.}
    \label{fig:more_ui}
\end{figure}

\begin{figure}[h]
    \centering
    \begin{subfigure}[t]{0.48\linewidth}
    \centering
        \includegraphics[width=\linewidth, height=3cm, keepaspectratio]{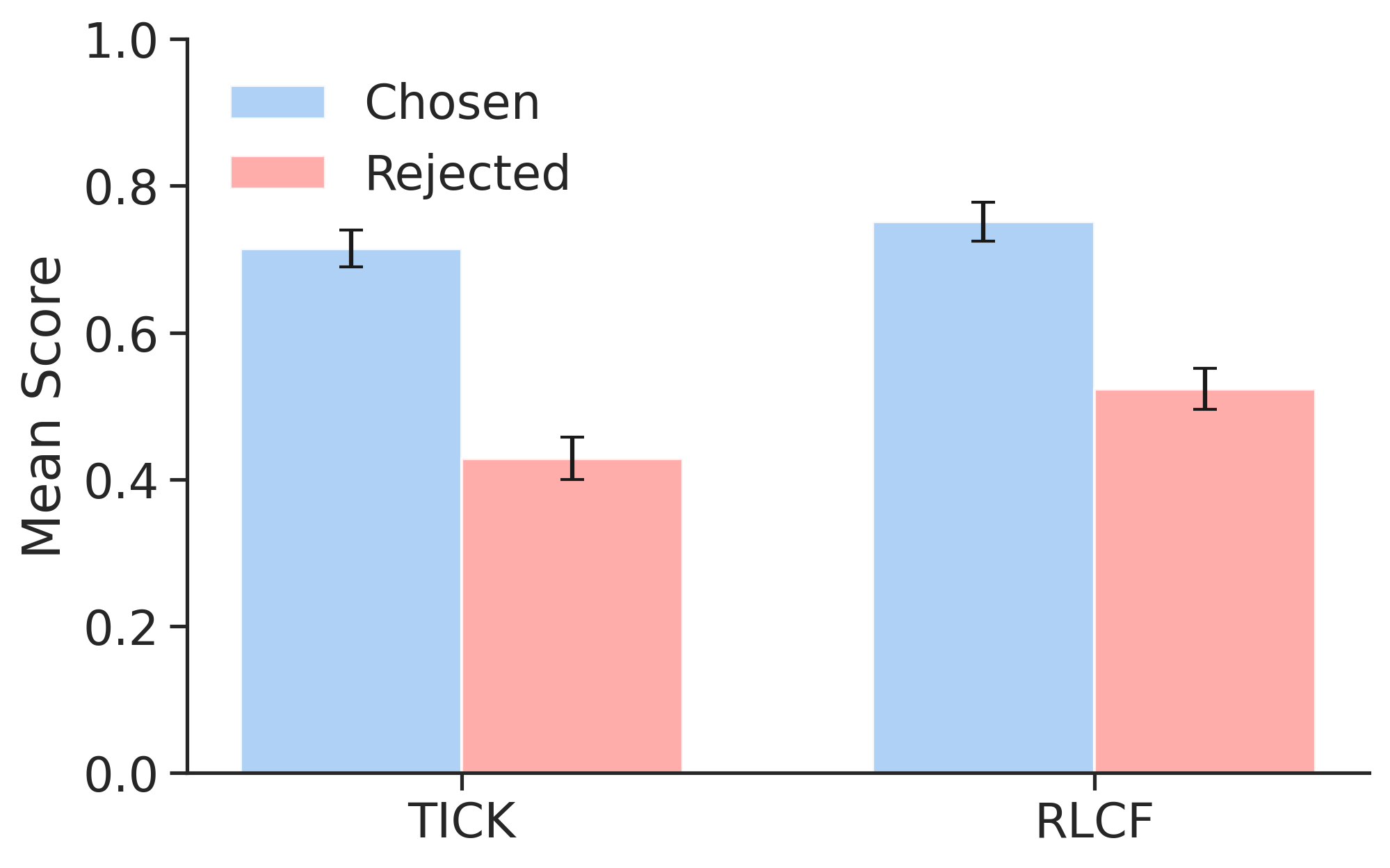}
    \caption{Mean checklist pass rates from \texttt{tick} and \texttt{rlcf} on RewardBench pairs.}
        \label{fig:rewardbench}
    \end{subfigure}
    \hfill
    \begin{subfigure}[t]{0.48\linewidth}
    \centering  
        \includegraphics[width=\linewidth, height=3.25cm, keepaspectratio]{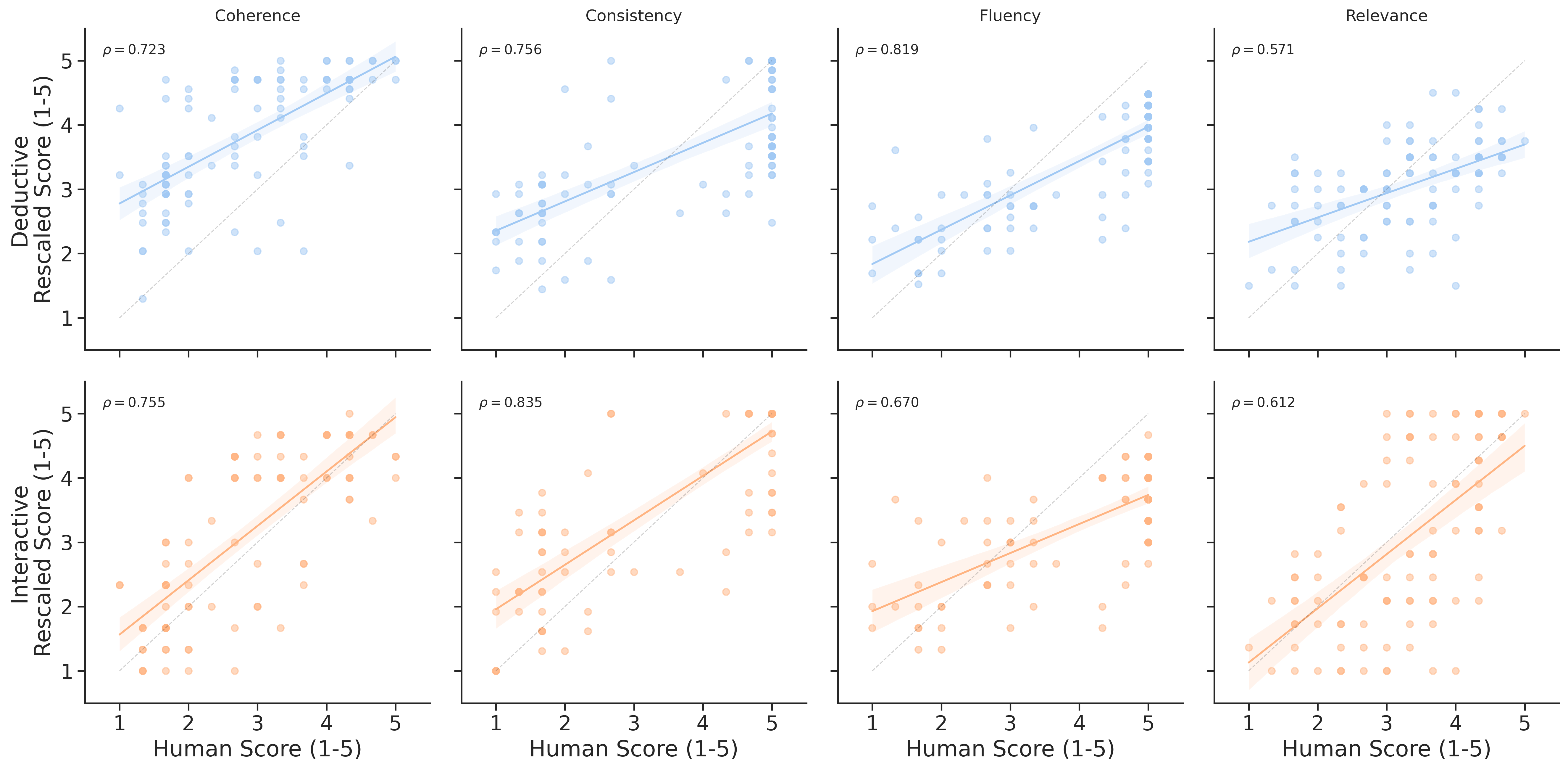}
    \caption{Rescaled checklist pass rates against human assigned scores on SummEval, across four dimensions.}
        \label{fig:summeval}
    \end{subfigure}
    \begin{subfigure}[t]{0.48\linewidth}
    \centering  
        \includegraphics[width=\linewidth, height=4cm, keepaspectratio]{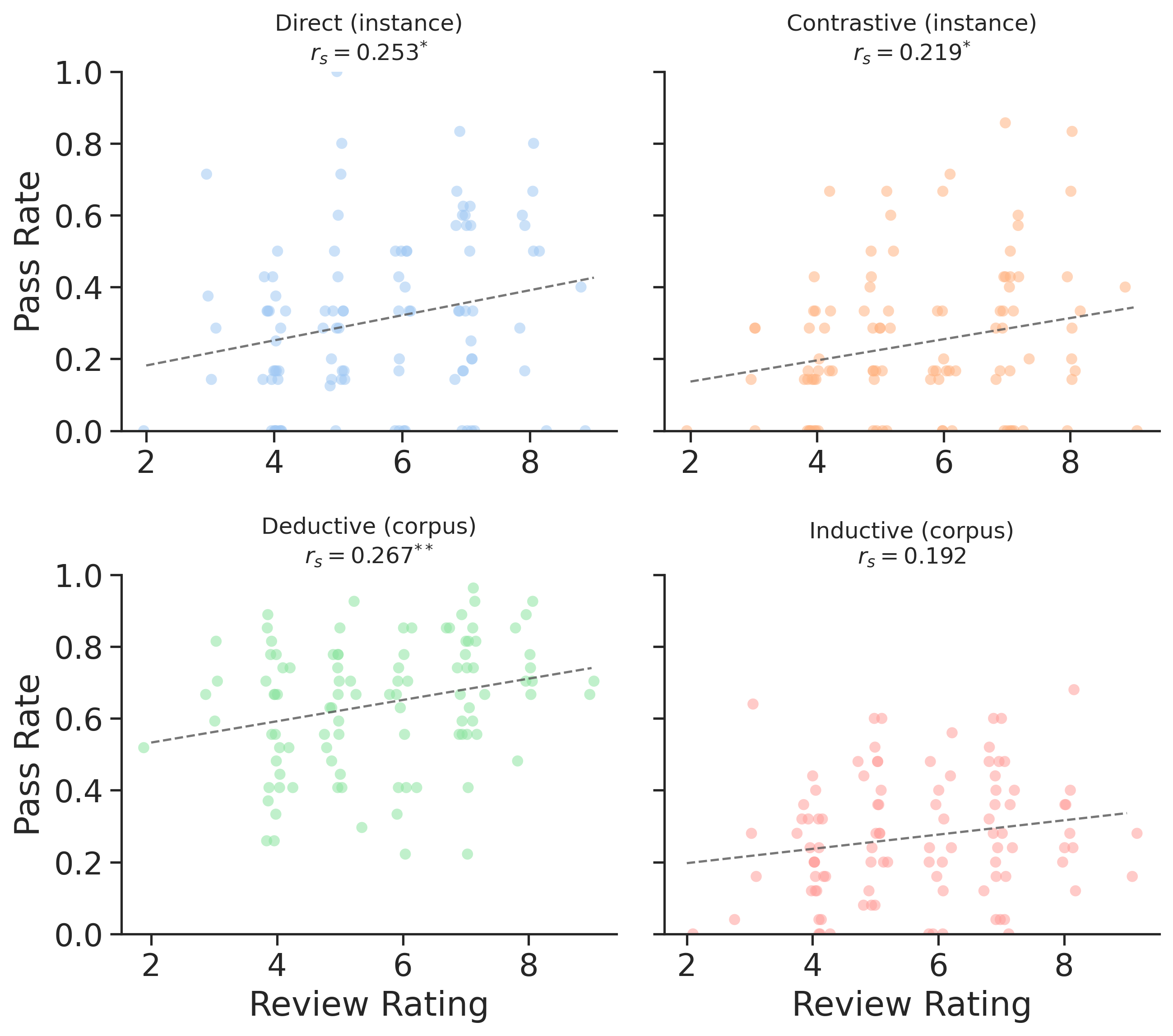}
    \caption{Checklist pass rate vs.\ reviewer rating on ICLR 2019 review--rebuttal pairs; dashed lines show linear fits. 
    }
        \label{fig:rating-correlation}
    \end{subfigure}
    \hfill
     \begin{subfigure}[t]{0.48\linewidth}
    \centering
        \includegraphics[width=\linewidth, height=3cm, keepaspectratio]{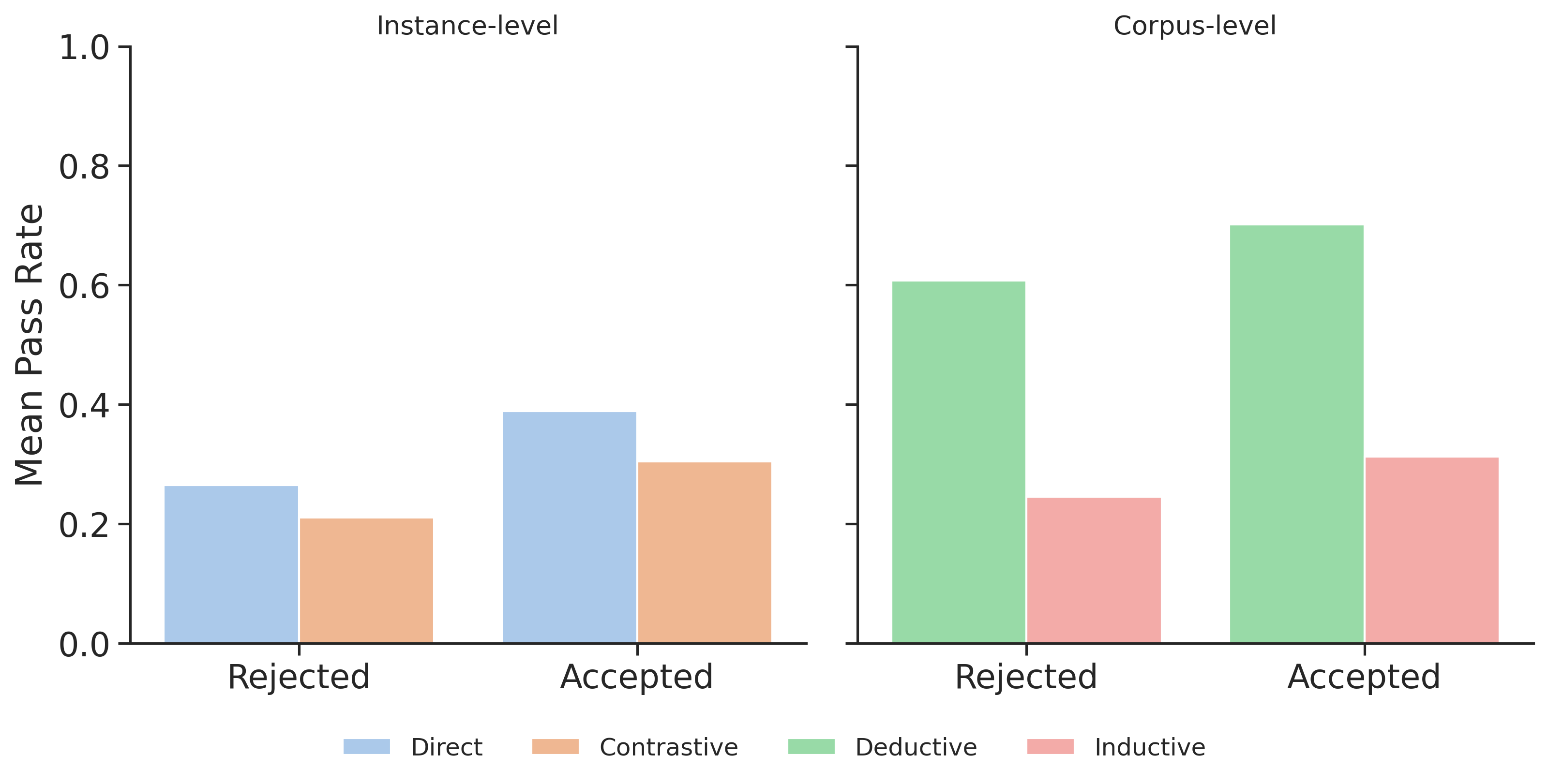}
    \caption{All four generators assign higher pass rates to rebuttals for accepted papers vs.\ rejected papers from ICLR 2019.}
        \label{fig:accepted-rejected}
    \end{subfigure}

    \caption{Additional results from validation and case studies.}
    \label{fig:more_results}
\end{figure}



\newpage
\section{Code Samples}
\label{sec:appendix}

\begin{python}[caption={Python API: basic usage}]
from autochecklist import pipeline

# Built-in pipeline: generate checklist + score in one call
pipe = pipeline("tick", generator_model="openai/gpt-5-mini", scorer_model="openai/gpt-5-mini")
result = pipe(input="Write a haiku about autumn.", target="Leaves fall gently down...")
print(f"Pass rate: {result.pass_rate:.0%}")

# Evaluation over a dataset
data = [{"input": "Write a haiku", "target": "Leaves fall..."}, ...]
result = pipe.run_batch(data, show_progress=True)
print(f"Macro pass rate: {result.macro_pass_rate:.0%}")
result.to_dataframe()  # or result.to_jsonl("results.jsonl")
\end{python}

\begin{python}[caption={Python API: custom prompts, providers, and composition}]
from autochecklist import DirectGenerator, ChecklistScorer, pipeline
from autochecklist.refiners import Deduplicator, Tagger
from pathlib import Path

# Custom generator prompt (inline or from a Markdown file)
gen = DirectGenerator(custom_prompt=Path("prompts/my_eval.md"), model="openai/gpt-5-mini")
checklist = gen.generate(input="Write a haiku about autumn.")

# Mix and match: chain refiners, then score with a different scorer
checklist = Deduplicator(model="openai/gpt-5-mini").refine(checklist)
checklist = Tagger(model="openai/gpt-5-mini").refine(checklist)
scorer = ChecklistScorer(mode="item", primary_metric="weighted", model="openai/gpt-5-mini")
score = scorer.score(checklist, target="Leaves fall gently down...")

# Or compose everything through the pipeline factory
pipe = pipeline("tick", generator_model="openai/gpt-5-mini",
                refiners=["deduplicator", "tagger"], scorer="weighted")

# Switch LLM provider: use a local vLLM server instead of OpenRouter
pipe = pipeline("tick", provider="vllm", base_url="http://localhost:8000/v1",
                generator_model="meta-llama/Llama-3-8B")

# Register and save custom pipelines for reuse
from autochecklist import register_custom_pipeline, save_pipeline_config
register_custom_pipeline("my_eval", generator_prompt=Path("prompts/my_eval.md"), scorer="weighted")
save_pipeline_config("my_eval", "my_eval.json")   # share as a portable JSON config

pipe = pipeline("my_eval", generator_model="openai/gpt-5-mini", scorer_model="openai/gpt-5-mini")
\end{python}

\end{document}